\documentclass[runningheads]{llncs}
\usepackage{graphicx}
\usepackage{comment}
\usepackage{amsmath,amssymb} 
\usepackage{color}

\begin{document}
\pagestyle{headings}
\mainmatter
\def\ECCVSubNumber{4997}  

\title{Attend and Segment: Attention Guided Active Semantic Segmentation} 

\titlerunning{Attend and Segment}
%
\author{Soroush Seifi\orcidID{0000-0002-4791-5350} \and
Tinne Tuytelaars\orcidID{0000-0003-3307-9723}}

\authorrunning{S. Seifi, T. Tuytelaars}
%
\institute{KU Leuven, Kasteelpark Arenberg 10, 3001 Leuven, Belgium
\email{\{FirstName.LastName\}@esat.kuleuven.be}\\
}
\maketitle

\begin{abstract}
In a dynamic environment, an agent with a limited field of view/resource cannot fully observe the scene before attempting to parse it. The deployment of common semantic segmentation architectures is not feasible in such settings. In this paper we propose a method to gradually segment a scene given a sequence of partial observations. The main idea is to refine an agent's understanding of the environment by attending the areas it is most uncertain about. Our method includes a self-supervised attention mechanism and a specialized architecture to maintain and exploit spatial memory maps for filling-in the unseen areas in the environment. The agent can select and attend an area while relying on the cues coming from the visited areas to hallucinate the other parts. We reach a mean pixel-wise accuracy of $78.1\%$, $80.9\%$ and $76.5\%$ on CityScapes, CamVid, and Kitti datasets by processing only $18\%$ of the image pixels (10 retina-like glimpses). We perform an ablation study on the number of glimpses, input image size and effectiveness of retina-like glimpses. We compare our method to several baselines and show that the optimal results are achieved by having access to a very low resolution view of the scene at the first timestep.

\end{abstract}

\begin{figure}[t]
 \begin{center}
     \includegraphics[width=\linewidth]{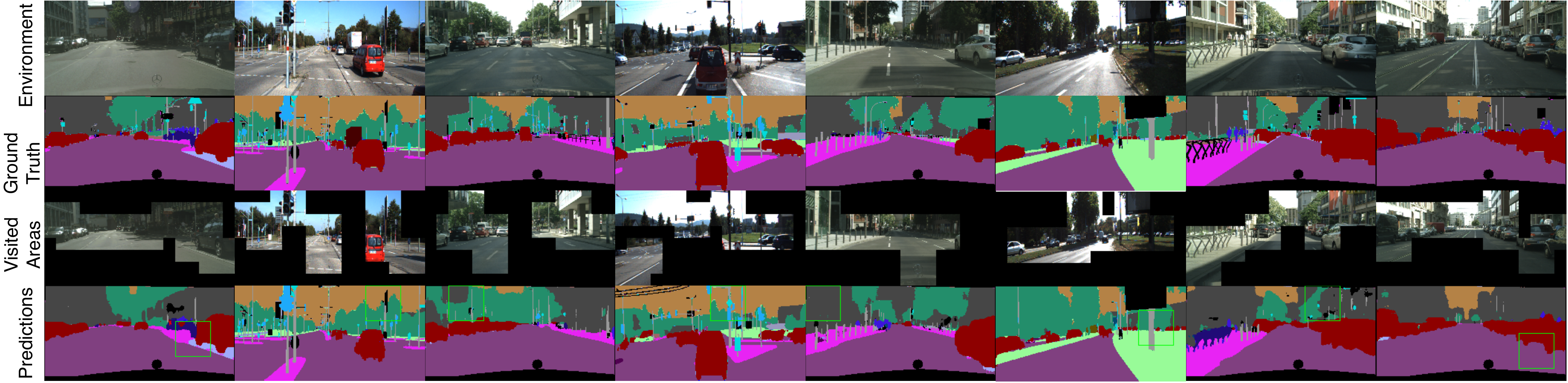}
   \end{center}
      \caption{Our model predicts a segmentation map for the full environment (last row) by attending 8 downscaled glimpses containing only 18\% of the pixels (third row).}
      \label{gt-seg}
\end{figure}

\section{Introduction}

Semantic segmentation has been extensively studied in the recent years due to its crucial role in many tasks such as autonomous driving, medical imaging, augmented reality etc. \cite{d1,d2,m1,m2}. Architectures such as FCN, U-Net, DeepLab etc. \cite{c1,c2,c3,c11} have pushed its accuracy further and further each year. All these architectures assume that the input is fully observable. They deploy deep layers of convolutional kernels on all input pixels to generate a segmentation mask. 

In contrast, in this paper we study the problem of parsing an environment with very low observability. 
We define an active agent with a highly limited camera bandwidth (less than $2\%$ of all input pixels) which cannot see the whole scene (input image) at once. Instead it can choose a very small part of it, called a `glimpse', to focus its attention on.
The agent has the freedom to change its viewing direction at each time step and take a new glimpse of the scene. However, depending on a pixel budget, it is limited in the number of glimpses it can see. After reaching this limit, the agent should output a segmentation map for the the whole scene including the unvisited areas. 

This setting is in line with previous works on `active visual exploration' such as \cite{c4,c5,c6} where an agent tries to explore, reconstruct and classify its environment after taking a series of glimpses. Inspired by those works, we take a step forward to solve an `active semantic segmentation' problem which: 1) is more practical compared to image reconstruction and 2) is more challenging compared to scene classification as there is a need to classify all visited and unvisited pixels. Furthermore we introduce a novel self-supervised attention mechanism which tells the agent where to look next without the need for reinforcement learning \cite{c4,c5} or supervision coming from the image reconstruction loss \cite{c6}.

Our agent is trained end-to-end, segments the visited glimpses and uses their extracted features to extrapolate and segment areas of the environment it has never seen before. We use specialized modules to segment the local neighbourhood of the glimpses and to exploit long-range dependencies between the visited pixels to segment the other unseen parts.

Our proposed method can be applied in scenarios where processing the whole scene in full resolution is not an option. This could be because 1) the agent's field of view is restricted and cannot capture the whole scene at once, 2) there is a limited bandwidth for data transmission between the agent and the processing unit, 3) processing all pixels from the scene in a sliding window fashion is redundant or impossible due to resource limitations, or 4) there is a need to process at least some parts in higher resolution.

We propose two solutions for such an agent: 1) Start from a random glimpse and intelligently choose the next few glimpses to segment the whole scene or 2) Start from  a (very) low resolution view of the whole scene and refine the segmentation by attending the areas with highest uncertainties. 
We show that the first method outperforms various baselines where the agent selects the next location based on a given heuristic while the second method can yield results comparable to processing the whole input at full resolution, for a fraction of the pixel budget.

Similar to the arguments in \cite{c4,c5,c6}, autonomous systems relying on high resolution 360$^\circ$ cameras could benefit the most from our architecture. However, due to lack of annotated segmentation datasets with 360$^\circ$ images we adapted standard benchmark datasets for semantic segmentation, namely CityScapes, Kitti and CamVid  \cite{d1,d2,d3}, to our setting. Figure \ref{gt-seg} illustrates the segmentations produced by our method after taking 8 retina-like glimpses on these datasets. We provide several baselines for our work along with an ablation study on the number of glimpses for each dataset. To the best of our knowledge, we are the first to tackle the problem of `active semantic segmentation' with very low observability.

 The remainder of this paper is organized as follows. Section 2 provides a literature review. Section 3 defines our method. In section 4 we provide our experimental results and we conclude the paper in section 5.


   
\section{Related Work}

\subsubsection{Semantic Segmentation}
Semantic segmentation is one of the key challenges towards understanding a scene for an autonomous agent \cite{c10}. Different methods and tricks have been proposed to solve this task relying on deep Convolutional Neural Networks (CNNs) \cite{c1,c2,c3,c11,c10,c13}. In this paper, we tackle the problem where an agent dynamically changes its viewing direction and receives partial observations from its environment. This agent is required to intelligently explore and segment its environment. Therefore, this study deviates from the common semantic segmentation architectures where the input is static and fully observable. Our work is close to \cite{c35} where an agent tries to segment an object in a video stream by looking at a specific part of each frame. However, in this work we produce a segmentation map for all input pixels for a static image.

\subsubsection{Active Vision}
Active vision gives the freedom to an autonomous agent to manipulate its sensors and choose the input data which it finds most useful for learning a task \cite{c14}. Such an agent might manipulate objects, move in an environment, change its viewing direction etc. \cite{c15,c16,c17,c18}. In this paper, we study the same active setting as \cite{c4,c5,c6} where an agent can decide where to look next in the scene (i.e. selecting a glimpse) with a goal of exploration. These studies evaluate their work on image reconstruction and scene classification. Such tasks demonstrate that the agent can potentially learn an attention policy and build a good representation of the environment with few glimpses. However, the practical use case for such an agent is not clear. Besides, the results from those works imply that the extrapolation beyond the seen glimpses in the image reconstruction case is mostly limited to filling in the unseen areas with uniform colors. Therefore, instead in this paper we tackle the active exploration problem for semantic segmentation where the agent needs to reason about the unseen areas and assign a semantic label to every pixel in the image. This allows focusing on the semantics, rather than the precise color or texture, which is difficult to predict. We believe such an agent is fundamentally more useful than the one solving an image reconstruction task. 

\subsubsection{Memory in Partially Observable Environments}
A critical challenge for an active agent in a partially observable environment is to understand the correlations and the spatial organization of the observations it receives. Many architectures combine LSTM layers with deep reinforcement learning to update their representation of the environment at each timestep \cite{c4,c5,c8,c9,c30,c31}. However, studies such as \cite{c6,c32,c33,c34} show that maintaining a spatial memory of the environment is more effective albeit being more expensive in terms of memory usage. In this study we use similar architectures to those proposed in \cite{c6,c35} and maintain the extracted features in spatial memory maps. These partially filled memory maps are exploited at each time step to segment the whole scene.

\subsubsection{Visual Attention}
We use the word `attention' to denote a mechanism for choosing the best possible location in the environment to attend next. This is different from those works in the literature where the attention mechanism weights the extracted features from the whole input according to their importance/relevance (a.k.a self-attention~\cite{c19,c20}, soft attention \cite{c7,c8,c21} or `global' attention \cite{c22}). Instead, this work is close to the hard attention mechanism defined in \cite{c8,c9,c35} where the information about the input is gathered sequentially by attending only a specific part of the input at each timestep. However, unlike the studies on hard attention, our attention mechanism does not rely on reinforcement learning, is differentiable and is trained with self-supervision. We take inspiration from \cite{c23} to derive an uncertainty value for each pixel in the predicted segmentation map. Consequently, the area with the highest uncertainty is attended to next.
\subsubsection{Image Generation and Out-painting}
Unlike various inpainting methods which reconstruct missing image regions based on their surrounding pixels \cite{c24,c25,c26}, image outpainting's purpose is to restore an image given only a part of it \cite{c27,c28,c29}. The active agent defined in \cite{c4,c5,c6} implicitly solves an outpainting problem. Such an agent should be able to exploit the spatial relationship of the visible areas to extrapolate and reconstruct the missing parts. Studies such as \cite{c4,c5,c36} incorporate the spatial information using explicit coordinates while \cite{c6,c35} maintain spatial memory maps for this purpose. In this study, we follow the later approach to extrapolate beyond the seen glimpses and assign a semantic label to each pixel in those regions.
\subsubsection{Retina Camera Technology}
Taking inspiration from the human's retina setting, our method benefits from the retina-like glimpses where the resolution changes spatially based on the distance to a point of interest \cite{c37}. This way the agent can use its pixel budget more efficiently. In this work we use common downscaling techniques to construct a retina-like glimpse. However, in practice, our method can be implemented on top of retina sensors introduced in \cite{c37,c38,c39} to visit the parts of the environment suggested by our attention mechanism without seeing and processing the other parts.



\section{Method}
\begin{figure}[t]
  \begin{center}
      \includegraphics[width=\linewidth]{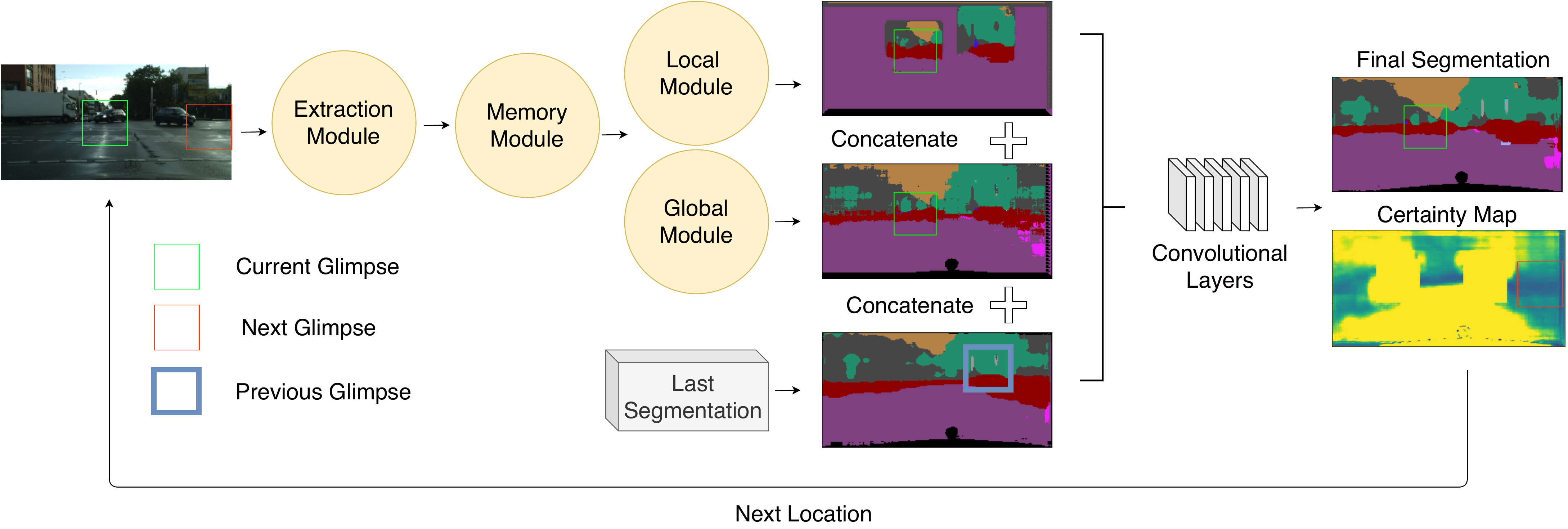}
   \end{center}
      \caption{Architecture Overview}
      \label{arch}
   \end{figure}

Our architecture consists of four main components. Figure \ref{arch} shows an overview of our architecture. The `Extraction Module' extracts features for each attended glimpse. The `Memory Module' gathers the features for all visited glimpses in spatial memory maps. The `Local Module' segments the attended regions and their neighborhood while the `Global Module' predicts a general layout of the whole scene. The final segmentation and uncertainty maps at each step are derived based on the outputs of the local and global modules and the final segmentation map from the previous step. The area with the highest uncertainty is selected as the next location for attendance. Figure \ref{arch} provides an overview of our architecture. In the following subsections we describe each module in more detail.

\subsection{Extraction Module}
\subsubsection{Retina Glimpses}
\begin{figure}
  \begin{center}
      \includegraphics[scale=0.5]{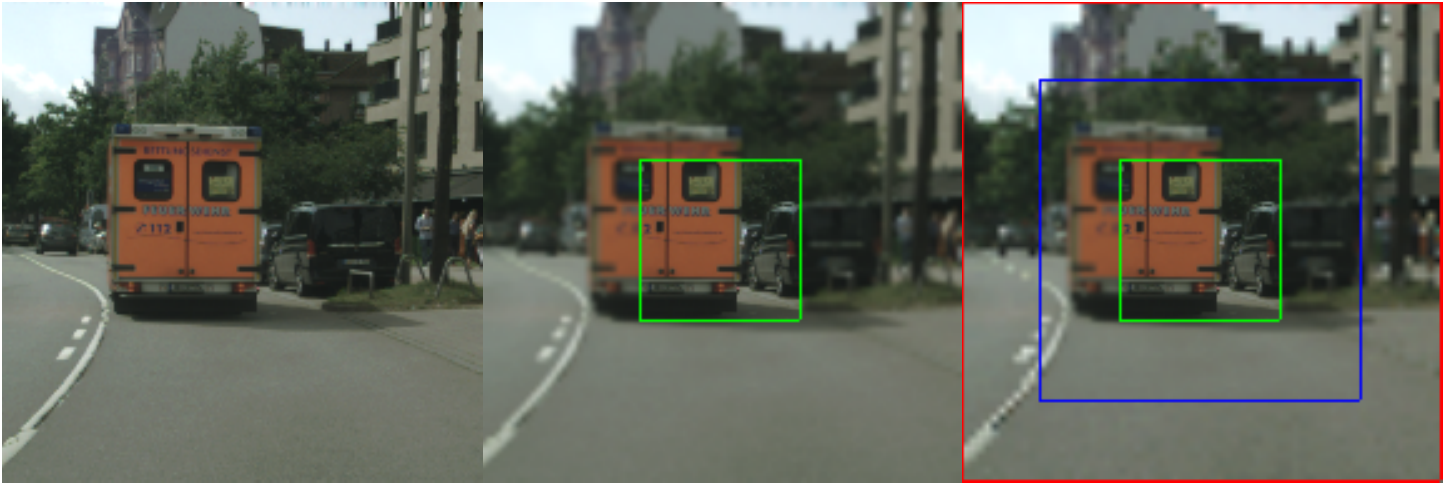}
   \end{center}
      \caption{Left to right: a glimpse in full-resolution, a retina glimpse with 2 scales and a retina glimpse with 3 scales. For a glimpse with size $48\times48$, there are 2304, 768 and 590 pixels from the original image in each one of these settings respectively. These images are only for illustration purpose and have a size of $96\times96$ rather than $48\times48$.}
      \label{glimpses}
   \end{figure}

The extraction module receives a glimpse which is scaled down on the areas that are located further from its center (`Retina-like glimpses' \cite{c6,c9}). This way the agent can use its pixel budget more efficiently. Figure \ref{glimpses} shows 3 different retina setting used in our experiments.
\subsubsection{Architecture}
\begin{figure}
  \begin{center}
      \includegraphics[scale=0.35]{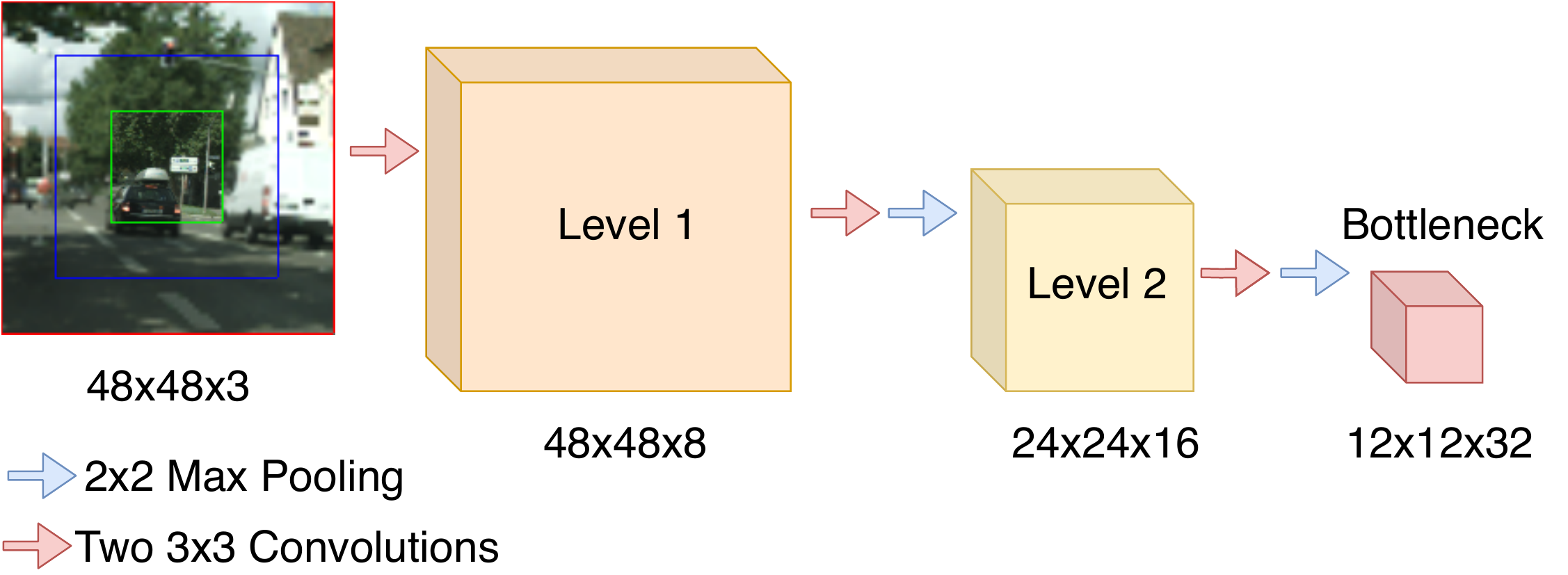}
   \end{center}
      \caption{Extraction module: The extracted features in each level of this encoder are stored for all glimpses by memory module.}
      \label{extraction}
   \end{figure}

This module uses a shallow stack of convolutional layers to extract features $F_t$ from the visited glimpse at time step $t$. Its architecture resembles the encoder part of U-net with only 32 channels for its bottleneck activations. Figure \ref{extraction} shows the architecture for this module.
   
\subsection{Memory Module}
\begin{figure}
  \begin{center}
      \includegraphics[width=\linewidth]{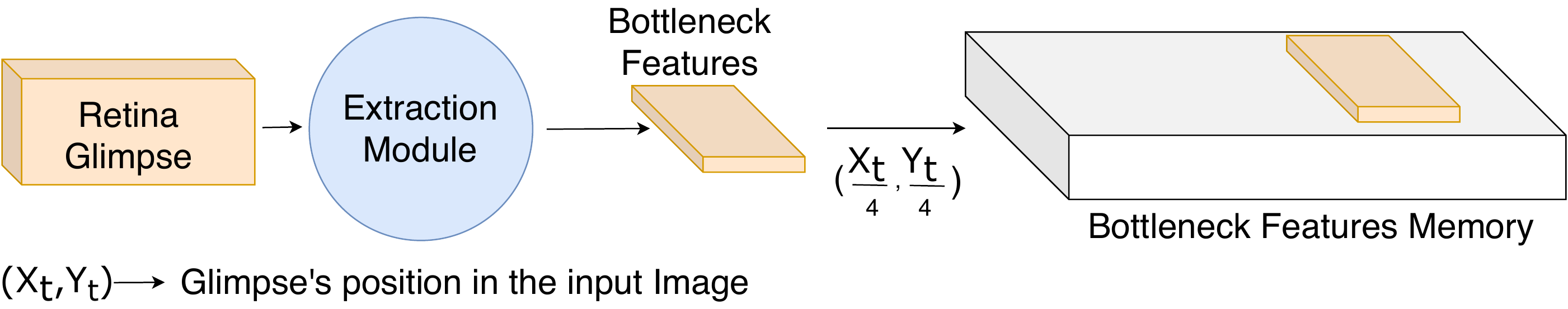}
   \end{center}
      \caption{Memory Module: Bottleneck features are stored in their corresponding spatial position in the memory.}
      \label{memory}
   \end{figure}

The memory module maintains 3 different matrices, one for each encoder level in Figure \ref{extraction}. We denote these matrices as `Level 1', `Level 2' (`intermediate' memories) and `Bottleneck' memory. In case that the agent visits all possible non-overlapping glimpses in the image, these matrices would contain the extracted features for the whole input image. Otherwise they are only partially filled with the information from the visited glimpses. In our setting, where the number of glimpses is limited, one can think of these memories as the representation for the whole input image after applying a dropout layer on top. This implicit drop out mechanism prevents the agent from overfitting to the data. Figure \ref{memory} illustrates the memory module for the `Bottleneck memory'; since bottleneck features are derived after two 2x2 pooling layers, their position in the feature memory is equal to glimpse's position in the image divided by 4. In case of overlap between two glimpses, these memories are updated with the features of the newest glimpse in the overlapping area.
\subsection{Local Module}
\begin{figure}
  \begin{center}
      \includegraphics[width=\linewidth]{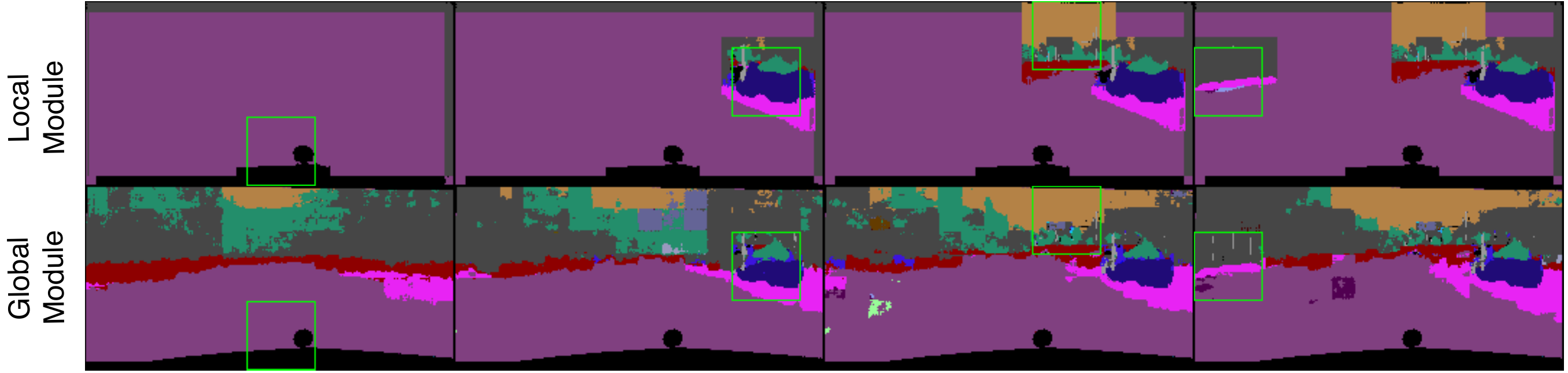}
   \end{center}
      \caption{Local module segments and expands the predictions for each glimpse while the global module predicts the general structure of the whole scene.}
      \label{fconve}
   \end{figure}

This module exploits the local correlations of the features in the memory to expand the segmentations for the visited glimpses. Since the convolutional kernels have a limited receptive field, these expansions remain local to each glimpse. At the same time, for two glimpses which are located close to each other it can benefit from the features from both glimpses to expand a larger area. Figure \ref{fconve} (top) illustrates this for 4 time-steps. 

The features in the `Bottleneck memory' are extracted using the encoder represented in figure \ref{extraction}. Consequently, we define a decoder architecture symmetrical with this encoder to generate the segmentations. The features in the `intermediate' memories are used as skip connections while decoding. The extraction and local module together define an architecture similar to U-net. However, the encoder extracts the features for each glimpse separately from the others while the decoder operates on a partially filled memory which contains the features for all glimpses visited until the current timestep. Figure \ref{local} illustrates the architecture of the local module. We denote the segmentation produced by this module at each step t as $L_t$ and measure its error $e_{L_t}$ using a binary cross-entropy loss.
\subsection{Global Module}
To complement the task of the local module, the global module exploits the long-range dependencies of the features in the memory and predicts the general structure of the scene. 

\begin{figure}
  \begin{center}
      \includegraphics[width=\linewidth]{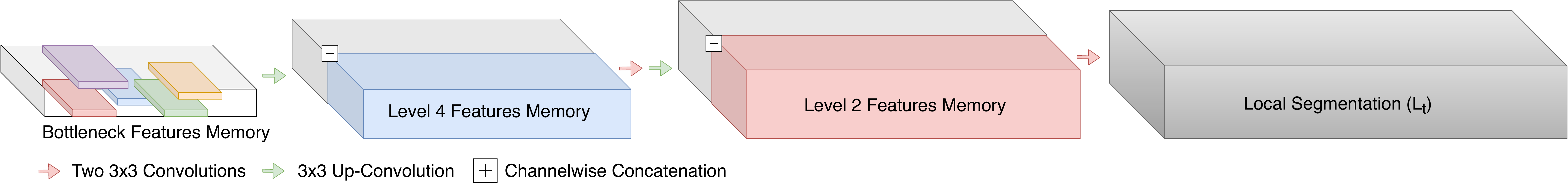}
   \end{center}
      \caption{Local Module's Architecture.}
      \label{local}
   \end{figure}

To achieve this, it compresses the `Bottleneck memory' with strided convolutions to 4 times smaller in each dimension (height, width and depth). Next, it deploys convolutional layers with a kernel size equal to the size of the compressed memory, thus taking into account all the features in the memory at once to predict a downscaled segmentation of the environment. This segmentation gets upscaled to the input's resolution with the help of `intermediate' memories and with a similar architecture to the one depicted in figure \ref{local} (though starting from a compressed bottleneck memory). Figure \ref{fconve} shows that the global module captures and mostly relies on the dataset's prior to hallucinate the unseen areas in the first steps. However, with more glimpses, its prediction changes towards the correct prediction of the structure of environment.

We denote the segmentation produced by this module at each step t as $G_t$ and again measure its error $e_{G_t}$ using a binary cross-entropy loss.

\subsection{Final Segmentation, Certainty and Attention}
At each step our architecture produces a segmentation map $S_t$ along with an extra channel $C_t$ as our certainty map. These maps are derived by concatenating the previous segmentation map $S_{t-1}$, the local segmentation $L_t$ and the global segmentation $G_t$ and using a series of convolution layers to combine them into a refined segmentation and a new certainty map. 

Inspired by the proposed method in \cite{c23} for learning the aleatoric and epistemic uncertainty meausures while optimizing the loss function, we define the loss for each module at step $t$ according to the equations \ref{eq:local}, \ref{eq:2} and \ref{eq:3}:
\begin{equation}
L_{L_t}=L_{L_{t-1}}+C_{t}\times e_{L_t}+U_{t}
\label{eq:local}
\end{equation}
\begin{equation}
L_{G_t}=L_{G_{t-1}}+C_{t}\times e_{G_{t}}+ U_{t}
\label{eq:2}\
\end{equation}
\begin{equation}
L_{S_t}=L_{S_{t-1}}+C_{t}\times e_{S_{t}}+ U_{t}
\label{eq:3}\
\end{equation}
$L_{L_0}$, $L_{G_0}$ and $L_{S_0}$ are initialized to zero. $C_{t}$ denotes the predicted certainty map at step $t$ while $U_t$ is a regularizer term to prevent minimizing the loss by setting $C_t$ to zero. We define $U_t$ as:
\begin{equation}
    U_t=\exp^{-C_t}
\end{equation}
$U_t$ measures the uncertainty for each pixel.
The agent learns to minimize $L_{L_t}$, $L_{G_t}$ and $L_{S_t}$ by assigning low values to $C_t$ (high values to $U_t$) in the areas where the loss is high (i.e. uncertain areas). Similarly, it assigns high values to $C_t$ (low values to $U_t$) for the areas with high certainty where the loss is low.

At step t, the optimizer minimizes the sum of the loss functions defined above. We denote this sum as $L_t$:
\begin{equation}
    L_t=L_{L_t}+L_{G_t}+L_{S_t}
    \label{eq:final}
\end{equation}

At the final stage of each step, the certainty map $C_t$ is divided into $16\times16$ non-overlapping patches and the patch with lowest sum (lowest certainty) is selected as the next location for attendance.
\section{Experiments}
We evaluate our method on the CityScapes, Kitti and CamVid datasets \cite{d1,d2,d3}. For the CityScapes dataset we report our results on the provided validation set while for the Kitti and CamVid datasets we set a random 20\% split of the data to validate our method.

\subsection{Retina Setting}
In a first experiment, we show our results for the 3 different retina settings depicted in figure \ref{glimpses}. In this figure, although all glimpses cover the same area, they differ in the number of pixels they process from the input image. Table \ref{retinas} compares the ratio of processed pixels to the input image size for different retina settings. Each glimpse covers a $48 \times 48$ patch of a $128\times 256$ input image (or $96\times96$ patch of a $256\times 512$ image). As is clear from this table, retina glimpses allow the agent to cover larger areas of the environment while efficiently using its pixel budget.

\begin{table}[h]
\centering
\resizebox{.6\textwidth}{!}{
\begin{tabular}{|l|p{2.5cm}|p{2.5cm}|p{2.5cm}|p{2.5cm}|}
\hline
\# Glimpses& Full resolution & 2 Scales & 3 Scales\\
\hline\hline
1 & 7.0 \% & 2.3\% & 1.8\% \\
2 & 14.0\% & 4.6\% & 3.6\%\\
3 & 21.0\% & 7.0\% & 5.4\%\\
4 & 28.1\% &9.3\% &7.2\%\\
5 & 35.1\% & 11.7\% & 9.0\% \\
6 &42.1\%& 14.0\% & 10.8\%\\
7 &49.2\% & 16.4\% & 12.6\%\\
8 &56.2\% & 18.7\% &14.4\%\\
9 &63.2\% & 21.0\% &16.2\%\\
10 &70.3\% & 23.4\% &18.0\%\\
\hline
\end{tabular}
}
\vspace{0.2cm}
\caption{Ratio of pixels in a glimpse to the image size for different retina settings.}
\label{retinas}
\end{table}

\begin{figure}
  \begin{center}
      \includegraphics[width=\linewidth]{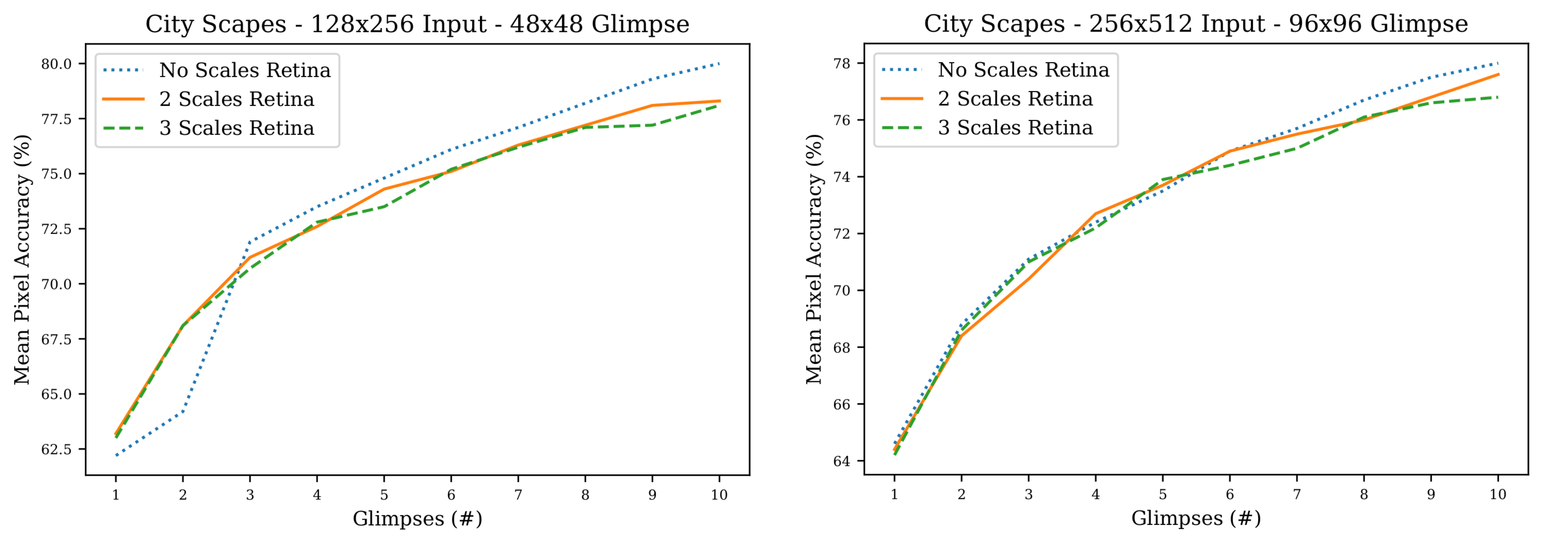}
   \end{center}
      \caption{Comparison of different retina settings' performance. 3-scales retina can perform equally well while using a much lower pixel budget.}
      \label{retinac}
   \end{figure}

Figure \ref{retinac} (Left) demonstrates the performance of our model for each retina setting. In these experiments we set the input image size to $128\times 256$ and each glimpse covers a $48\times48$ patch of the input. Similarly, the right part of this figure summarises the experiments where the input image size is $256\times512$ and each glimpse covers a $96\times96$ area of the input (ratios remain consistent with table \ref{retinas}).

Table \ref{retinas} and figure \ref{retinac} imply that the agent can use its pixel budget most efficiently using the 3-scales retina setting. An agent with a pixel budget of $18\%$ can achieve an accuracy of $78.1\%$ with 3 scales. With the same pixel budget, the 2-scales glimpse and full resolution glimpse cover a smaller area of the input image and thus their accuracy decreases to less than $77.2\%$ and $71.9\%$ respectively.

Furthermore, a comparison of the left and the right part of figure \ref{retinac} implies that if we maintain the ratio for the glimpse's coverage according to the input size, our method achieves similar results. Therefore, we evaluate the rest of our experiments in this paper using the $128\times256$ input size and a 3-scales retina with a coverage of $48\times48$ pixels. Table \ref{citykiti} reports the results for Cityscapes, Camvid and Kitti datasets in such settings.

\begin{table}[h]
\centering
\begin{tabular}{|l|p{2cm}|p{2cm}|p{2cm}|p{2cm}|}
\hline
Glimpses& CityScapes & Camvid & Kitti\\
\hline\hline
1 & 63\% & 68.2\% & 64.3\% \\
\hline
2 & 68.1\% & 73.0\% & 69.6\%\\
\hline
3 & 70.7\% & 75.3\% & 72.1\%\\
\hline
4 & 72.8\% &77.8\% &72.4\%\\
\hline
5 & 73.5\% & 78.5\% & 73.2\% \\
\hline
6 &75.2\%& 78.9\% & 74.9\%\\
\hline
7 &76.2\% & 79.8\% & 75.1\%\\
\hline
8 &77.1\% & 80.4\% &75.3\%\\
\hline
9 &77.2\% & 80.6\% &76.0\%\\
\hline
10 &78.1\% & 80.9\% &76.1\%\\
\hline
\end{tabular}
\caption{Mean Pixel Accuracy for each dataset for different number of glimpses.}
\label{citykiti}
\end{table}
\subsection{Baselines}
In this section we evaluate our attention mechanism using different baselines. We compare against a `random agent' which selects the next glimpse's location by randomly sampling from the input locations. Next, we consider the fact that the images in the datasets with road scenes are captured through a dashboard camera. In this case, salient parts of the image typically lie somewhere near the horizon. Consequently, we compare our method against a `Horizon agent' where it can only look at the uncertain areas in the middle rows of the image. Finally, we compare our method against a `Restricted Movement agent' that looks at positions nearby to the current glimpse in the next step. This baseline is in line with the setting in previous literature on image reconstruction \cite{c4,c5}. It evaluates our attention mechanism's exploratory performance and our method's ability to correlate glimpses coming from far spatial locations.

\begin{figure}
  \begin{center}
      \includegraphics[width=0.5\linewidth]{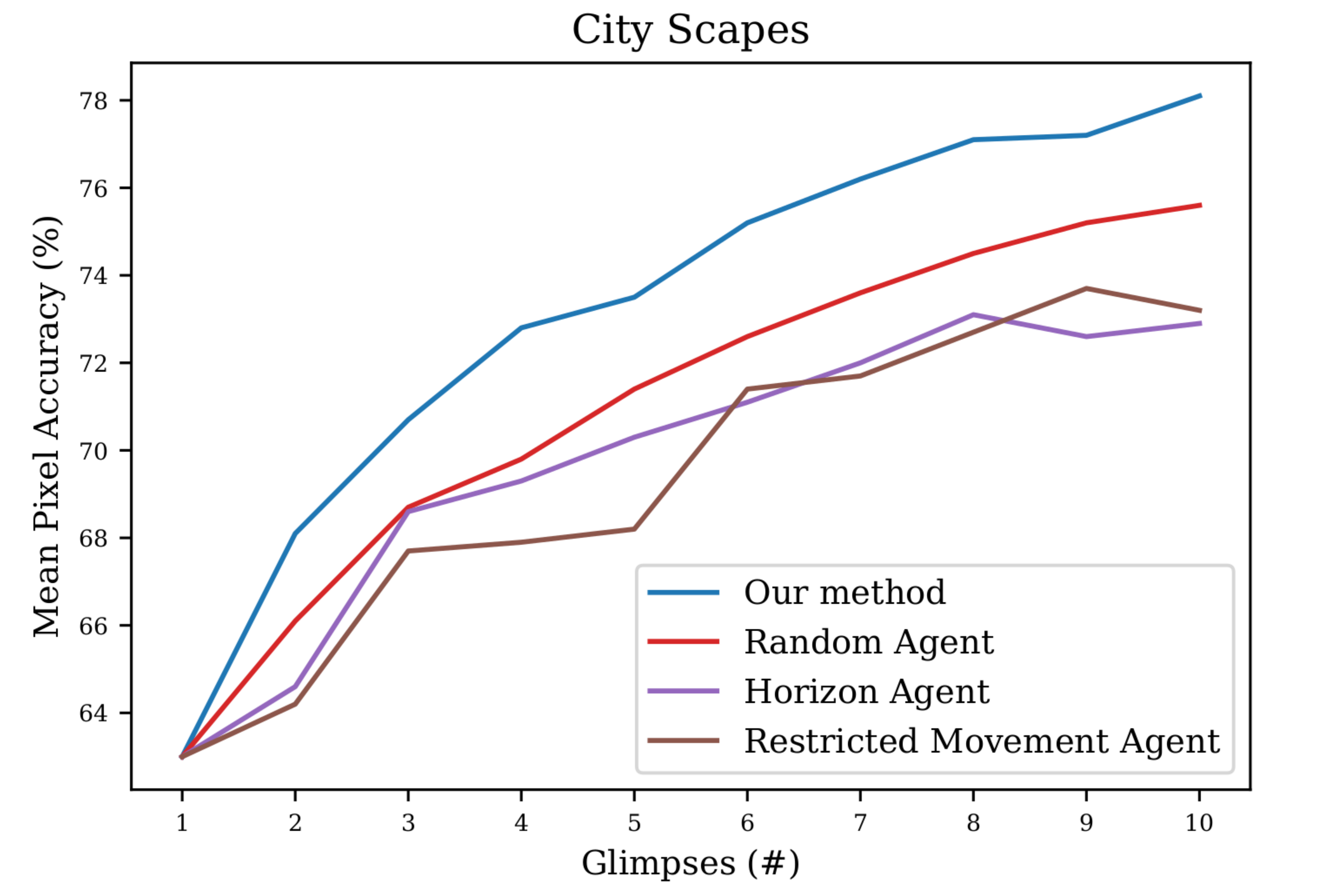}
   \end{center}
      \caption{Comparison against baselines.}
  \label{baselines}
\end{figure}

\begin{figure}
  \begin{center}
      \includegraphics[width=\linewidth]{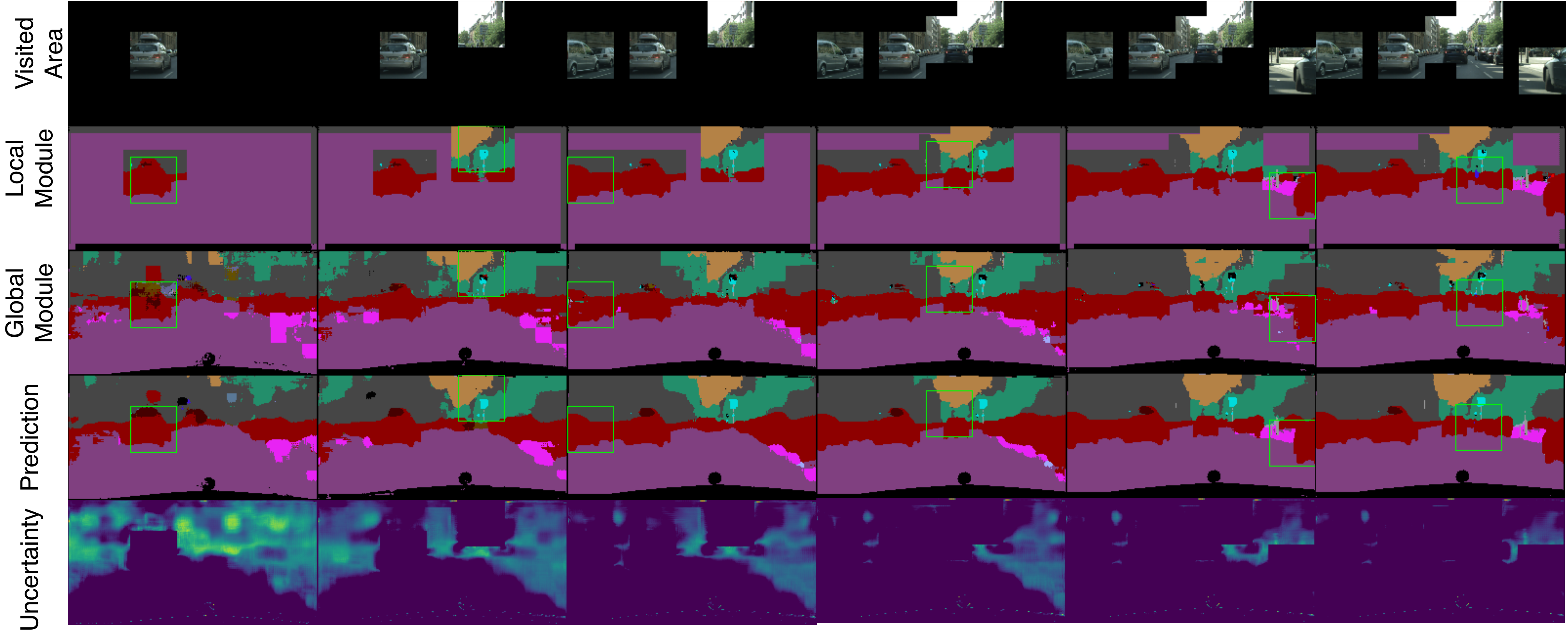}
   \end{center}
      \caption{The glimpse-only agent refines its predictions by attending the most uncertain areas. The local module expands the segmentations for the visited areas. The global module predicts the general layout of the environment. The final segmentation is derived by combining the last step's segmentation (initialized to zero) and the local and global modules' segmentations.}
      \label{steps}
  \end{figure}

Figure \ref{baselines} summarises our results on CityScapes dataset (See suplementary material for Camvid and Kitti.)
Results presented in figure \ref{baselines} suggest that remaining local to the horizon or the visited regions of the image forces the agent to hallucinate larger parts of the environment thus making the task more difficult. Furthermore, overlapping glimpses which are more likely to occur for the horizon and restricted movement agents can potentially waste a part of the agent's pixel budget without adding much information for the segmentation. Therefore, solving this task requires a more sophisticated strategy for exploration of the input rather than scan of the nearby locations. Finally, the comparison between our method and the random agent shows the effectiveness of our proposed attention/uncertainty prediction. Figure \ref{steps} confirms this by illustrating the output of the glimpse-only agent's modules for 6 time-steps. While remaining uncertain about most parts of the environment after the first glimpse, the agent imagines itself to be in a road with cars to its side. By taking the next glimpse above the horizon it predicts the general structure of the buildings and trees surrounding the road. In the few next steps it attends the areas along the horizon which contain more details that the agent is uncertain about.

\subsection{Glimpse-only, Hybrid and Scale-only agents}
In this section, we propose an extension of our proposed method which can achieve higher accuracy with smaller number of glimpses in case it is allowed to capture the whole scene at once at a low resolution. To evaluate this, we define three agents for the experiments in this section: 1) Glimpse-only agent: Similar to the previous experiments, the agent cannot capture the whole scene at once. It takes the first glimpse randomly and relies on the attention mechanism to select the attended areas in the next steps. 2) Hybrid agent: The agent can capture the whole scene but cannot process all pixels. It dedicates a part of its pixel budget to see the whole scene in low resolution. This helps the agent to capture the general structure of the environment and use its remaining pixel budget to refine its segmentation by attending the uncertain areas. For this setting we experimented with an agent which scales down the input to 32x32 (see supplementary materials for 16x8), which corresponds to almost 2 retina glimpses with 3 scales. 3) Scale-only agent: The agent `must' scale down the whole scene to its pixel budget. In this case, it does not take any glimpses and only relies on the scaled down view of the input. We define this agent as a baseline for the hybrid agent. The hybrid and scale-only agents use an architecture similar to the extraction module to encode the downscaled input. These features are decoded to a segmentation map using a symmetrical architecture to the extraction module. This would resemble a shallow U-net architecture. The scale-only agent upscales its segmentation to the input's resolution with bilinear interpolation.

\begin{figure}
  \begin{center}
      \includegraphics[width=\linewidth]{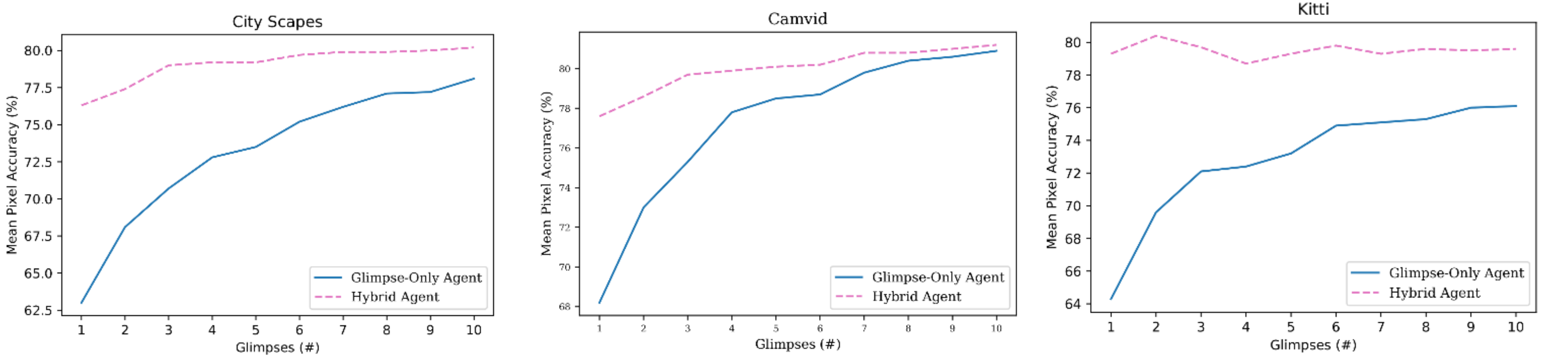}
   \end{center}
      \caption{Our method's performance for different number of glimpses. The gap between the glimpse-only and the hybrid agent decreases for higher number of glimpses.}
      \label{hybrids}
\end{figure}

Figure \ref{hybrids} and table \ref{dscaled} summarise our results for the agents defined above. 
As is clear from Figure \ref{hybrids}, the hybrid agent outperforms the glimpse-only one. However, the performance gap between these two agents decreases with the number of glimpses. For smaller number of glimpses the glimpse-only agent needs to hallucinate larger parts of the environment while the hybrid agent can rely on the downscaled input to fill-in the missing parts. Another interesting property for the hybrid agent is that it can achieve optimal results in much smaller number of steps (e.g.~2 glimpses in case of Kitti.)

Finally, a comparison between table \ref{dscaled} and figure \ref{hybrids} suggests that the glimpse-only agent performs favorably compared to the scale-only agent given the same pixel budget. However, in most cases the hybrid agent performs the best. This is due to the fact that such agent can decide which areas to attend in full resolution while its scaled down view of the scene is sufficient for parsing the other areas.

\begin{table}
{\footnotesize
\begin{center}
\begin{tabular}{|l|c|c|c|c|}
\hline
Scales& Glimpse Budget &CityScapes&Camvid&Kitti\\
\hline\hline
1 ($128\times 256$) (Full)&$\approx 56$&80.7 & 81.3 &81.7\\
1/4 ($64 \times128$)&$\approx 14$&80.4 & 80.9&80.4\\
1/16 ($32 \times 64$)&$\approx 4$& 78.9& 79.4&75.5\\
\hline
\end{tabular}
\end{center}
\caption{Scale-only agent; segmentation results by scaling down the input. Second column denotes the number of possible retina-like glimpses given the pixel budget for each experiment.}
\label{dscaled}
}
\end{table}

\subsection{IOU Evaluation}
In this section we compare the Mean IOU accuracy of the glimpse-only agent with 10 glimpses to the accuracy of an architecture similar to U-net (with 256 channels at its bottleneck) working on full $128\times256$ images from the CityScapes dataset. Table \ref{iou} compares our results for different categories in this dataset. For this evaluation all segmentations are bilinearly upscaled to the raw input image size of ($1024\times2048$).

\begin{table}[h]
\begin{center}
\begin{tabular}{|l|p{2cm}|p{2cm}|p{2cm}|}
\hline
Category& Our Method & U-net\\
\hline\hline
Flat & 0.907 & 0.938 \\
\hline
Construction & 0.641 & 0.746\\
\hline
Object & 0.046 & 0.138 \\
\hline
Nature & 0.647 &0.808\\
\hline
Sky & 0.503  & 0.809\\
\hline
Human & 0.216 & 0.006\\
\hline
Vehicle & 0.599 & 0.798\\
\hline
\hline
Average & 0.508 & 0.590\\
\hline
\end{tabular}
\end{center}
\caption{Mean IOU comparison on CityScapes dataset. Our method using only 18\% of the pixels in the image comes relatively close to U-net which observes the full image.}
\label{iou}
\end{table}

Our method compares well to an architecture working on the full image taking into account that our approach only processes $18\%$ of the input pixels. The most difficult category for our method is `Object'. In a partial view of an environment it is easy to miss small objects such as traffic signs and poles. Therefore it would be a difficult task for our method to hallucinate such objects lying in the unseen regions of the environment.

\section{Conclusion}
By taking inspiration from the recent works on active visual exploration \cite{c4,c5,c6}, in this study we tackled the problem of semantic segmentation with partial observability. In this scenario an agent with limited field of view and computational resources needs to understand the scene. Given a limited budget in terms of the number of pixels that can be processed, such an agent should look at the most informative parts of an environment to segment it in whole. We proposed a self-supervised attention mechanism to guide the agent on deciding where to attend next. The agent uses spatial memory maps and exploits the correlations among the visited areas in the memory in order to hallucinate the unseen parts of the environment. Moreover, we introduced a two-stream architecture, with one stream specialized on the local information and the other working on the global cues. We demonstrated that our model performs favorably in comparison to a solution obtained by scaling down the input to the pixel budget. Finally, our experiments indicated that an agent which combines a scaled down segmentation of the whole environment with the proposed attention mechanism performs the best. 

In the future, we would investigate datasets with less prior knowledge consisting of various scene categories such as ADE20k \cite{c40}. Next, having in mind that consecutive frames in a video stream share most of their content, we would look into a video segmentation problem with partial observability.

\paragraph{\textbf{Acknowledgment}}
This work was supported by the FWO SBO project Omnidrone \footnote{https://www.omnidrone720.com/}.



\clearpage
%
%
\bibliographystyle{unsrt}
\bibliography{egbib}
\end{document}